\documentclass{article} 
\usepackage{iclr2019_conference,times}

\usepackage{amsmath,amsfonts,bm}

\def\eqref#1{equation~\ref{#1}}

\def\1{\bm{1}}

\def\vtheta{{\bm{\theta}}}
\def\veta{{\bm{{\eta}}}}

\def\vx{{\bm{x}}}

\DeclareMathAlphabet{\mathsfit}{\encodingdefault}{\sfdefault}{m}{sl}
\SetMathAlphabet{\mathsfit}{bold}{\encodingdefault}{\sfdefault}{bx}{n}

\def\sS{{\mathbb{S}}}

\newcommand{\mtr}{m_{\rm{train}}}
\newcommand{\mte}{m_{\rm{test}}}
\newcommand{\padv}{p_{\rm{adv}}}

\newcommand{\pdata}{p_{\rm{data}}}

\newcommand{\pmodel}{p_{\rm{model}}}

\DeclareMathOperator*{\argmax}{arg\,max}

\usepackage{hyperref}
\usepackage{graphicx}
\usepackage{url}
\usepackage{tabu}

\title{A Research Agenda: Dynamic Models to Defend Against Correlated Attacks}

\author{Ian Goodfellow \thanks{\url{www.iangoodfellow.com}} \\
Google Brain\\
\texttt{ian-academic@mailfence.com}
}

\iclrfinalcopy 
\begin{document}

\maketitle

\begin{abstract}
In this article I describe a research agenda for securing machine learning
models against adversarial inputs at test time.
This article does not present results but instead shares some of my thoughts
about where I think that the field needs to go.
Modern machine learning works very well on I.I.D. data: data for which each
example is drawn {\em independently} and for which the distribution generating
each example is {\em identical}.
When these assumptions are relaxed, modern machine learning can perform
very poorly.
When machine learning is used in contexts where security is a concern,
it is desirable to design models that perform well even when the input
is designed by a malicious adversary.
So far most research in this direction has focused on an adversary who violates
the {\em identical} assumption, and imposes some kind of restricted worst-case
distribution shift.
I argue that machine learning security researchers should also address
the problem of relaxing the {\em independence} assumption and that current
strategies designed for robustness to distribution shift will not do so.
I recommend {\em dynamic models} that change each time they are run as a potential
solution path to this problem, and show an example of a simple attack
using correlated data that can be mitigated by a simple dynamic defense.
This is not intended as a real-world security measure, but as a recommendation
to explore this research direction and develop more realistic defenses.
\end{abstract}

\section*{Notation and definitions}
\label{sec:notation}

\begin{tabular}{p{2in}|l}
$\vx$ & A train or test example or a sample from the model \\
\hline
  $y$ & The true label for an example \\
\hline
$\vtheta$ & The parameter vector of the model \\
\hline
$\mtr$ & Number of training examples  \\
\hline
$\mte$ & Number of test examples \\
\hline
  $k$ & The number of classes \\
  \hline
  $r$ & The error rate of the classifier on the naturally occurring test set \\
  \hline
  $\tilde{r}$ & The error rate under the test set attack \\
  \hline
$\pmodel( \vx ; \vtheta) $ & The probability distribution over output classes learned by the model \\
\hline
  $\tilde{\vx}$  & An adversarial example\\
  \hline
  $f(\vx,\veta)$ & A transformation of $\vx$ controlled transformation parameters $\veta$.\\
  \hline
  $\sS$  & The set of allowable values of $\veta$\\
  \hline
  $L(\vx, y)$ & The loss incurred by the model while classifying input $\vx$ with label $y$.\\
  \hline
  $\epsilon$ & The constraint on adversarial perturbation size
\end{tabular}

\section{Introduction}

Machine learning is now a working technology and produces predictions that are correct
most of the time for many different tasks \citep{taigman2014deepface,he2015delving,wu2016google,amodei2016deep}.
In general, these tasks use ``naturally occurring'' data as opposed to data produced
by an adversary who intentionally tries to fool the model.
When an adversary tries to fool the model, the adversary generally succeeds
(see \citet{evaluation2019} for a recent evaluation guidelines
explaining how to implement attacks that almost always suceed).
The adversary can not only cause the model to make an incorrect prediction (an
{\em untargeted attack}, e.g. fail to recognize that a photo of a dog is dog)
but can also cause the model to make a specific mistake (a {\em targeted} attack,
e.g. cause the model to think the photo of a dog is specifically an airplane, or
some other class chosen by the attacker in advance).

In general, the reason that machine learning performs so much worse under attack
is that modern machine learning mostly relies on the {\em i.i.d. assumptions}.
We can think of most machine learning algorithms as using a training set to learn
the model with the goal of maximizing performance on a test set of examples.
Under the i.i.d. assumptions, the train and test  examples are all generated {\em independently}
from an {\em identical} distribution.
That is, each example is drawn independently from some distribution $\pdata$
that remains the same throughout the entire train and test generation process.

When a machine learning system is used in a setting where security is a concern,
the i.i.d. assumptions are usually no longer valid.
In this article I will focus on test-time attacks against the input of the model.
In this setting, the attacker supplies some or all of the test examples.
The attacker can draw test samples from a new distribution $\padv$ rather than
from $\pdata$.
This makes the challenge for the model much greater: instead of statistically
generalizing from training examples to new examples from the same distribution,
the model must also generalize to a different distribution.
In most work on adversarial examples so far, the test examples are still generated
indepedently from one another, and $\padv$ remains similar to $\pdata$, in the
sense that a sample from $\padv$ is generated by generating a sample from $\pdata$
and then modifying it slightly.
In this article, I argue that machine learning security researchers should additionally
study the setting where test set examples are not independent.
One attack strategy in this setting is to use early examples sampled randomly
from some simple distribution (such as $\pdata$) until a mistake is found, and
then late examples are all copies of a known mistaken point.
For modern models, these attacks can drive the error rate after the first mistake
is found to 100\%.

Modern machine learning models typically have two distinct phases of existence:
first, they go through a {\em training} stage, in which the parameters are adapted to
fit then training set, then the parameters are frozen and the model is deployed to
the {\em test } / {\em production } / {\em inference } / {\em serving} stage where
the same parameters are used to make predictions on new data indefinitely.
This split between the training stage and inference stage has been useful because
training updates are not very reliable and can often ruin the model.
Currently it is possible to rigorously evaluate a model and determine that its predictions
are reliable, then deploy the model to run in inference mode, but it is not yet feasible
to verify ahead of time that a particular training algorithm will produce reliable updates indefinitely.
Unfortunately, I believe that this separation of dynamic training and static inference
must end in order to defend against correlated test-time attacks.
To avoid the attacker, I believe that the model must become a moving target that continually changes,
even after it has been deployed.

\section{Attacks that violate the {\em identical} assumption}

Most adversarial examples attacks so far work by violating the {\em identical} assumption.
Models are designed assuming that all the test data is drawn from $\pdata$ (i.e., all the train
and test data is drawn from an identical distribution) but instead
the adversary supplies examples sampled from $\padv$.
On top of this, most work on adversarial examples constrains $\padv$ to be highly similar
to $\pdata$, in the sense that samples $\tilde{v}$ from $\padv$ are generated by slightly modifying
samples $\vx$ from $\pdata$.
A common attack strategy is to modify a sample $\vx$ from $\pdata$ by replacing it with
$f(\vx, \veta)$, where $f$ is some transformation of $\vx$ that preserves semantics when
$\veta$ is constrained to some set $\sS$ of allowable values. A common choice is
$f(\vx, \veta) = \vx + \veta$ for $\| \veta \|_\infty < \epsilon$ for some small value of
$\epsilon$.
Within these similarity constraints, adversarial examples are often chosen to maximize the loss
$L(\vx, y)$ incurred by the model.
The generation process is thus:
\[ \vx, y \sim \pdata \]
\[ \tilde{\vx} = \max_{\veta \in \sS} L(f(\vx, \veta), y). \]
The maximization operation is generally approximated with some kind of reasonably cheap optimization
algorithm and is often stochastic \citep{Goodfellow-2014-adversarial,kurakin2016physical,warde201611,madry2017towards}.
Altogether this generation process implicitly defines a distribution $\padv$.

Defenses against adversarial examples are often evaluated in terms of their error rate
across an adversarially perturbed version of the naturally occurring test set.
This metric was introduced by \citep{Goodfellow-2014-adversarial} and is still recommended, e.g.
by \citet{evaluation2019}.
This metric measures expected error in terms of starting points sampled independently from $\pdata$
and worst case error in terms of the $\veta$ transformation parameters applied at each starting point.
The metric is thus not truly worst case, because the choice of starting point is treated as
random and naturally occuring rather than adversarial.

\section{Attacks that violate the {\em independent} assumption}

In threat models where the attacker could realistically choose the ``starting point,'' the true worst
case metric.

This setting has previously been described by
a research agenda talk \citep{dark_arts},
a review paper \citep{gilmer2018motivating},
and is the basis for a contest \citep{brown_contest}.
Compared to \citet{dark_arts}, this article focuses much more on the problem of
non-independent test time inputs.
\citet{gilmer2018motivating} (Sec 4.8) describe correlated input attacks but do not advocate researching
defenses against them but instead advocate reducing the error rate on naturally occurring test sets
or reducing the total volume of errors made by the model.
This article argues that such approaches do not provide a defense against a correlated input attacker
and that defense mechanisms other than reduced error rate are necessary.
\citet{brown_contest} introduce a contest whose eventual winner will most likely need to develop
defenses against correlated input attacks, but the report introducing the contest does not include
suggestions for possible solutions.
This article focuses primarily on speculation about future directions that may lead to a solution.

\section{Threat modeling: when can an attacker correlate test time inputs?}

Security research should generally include a threat model: what are the attackers goals and capabilities?
In this case, we should consider which kinds of attackers are capable of mounting a test time input
attack.

At the same time, much of the point of machine learning research is that machine learning has the potential
to be fairly general, and applied to many application areas.
In this article, I don't attempt to exhaustively threat model all application areas, but just give some
brief examples to show how generic machine learning principles interact with threat modeling in specific
application areas.

Consider one potential attack vector: evading facial recognition by wearing glasses with patterns on the frames
designed to fool the face recognition system \citep{sharif2016accessorize}.
Suppose that when a person physically approaches the entrance to the facility, a guard
photographs them and the face recognition system returns an estimate of that person's specific
identity.
Suppose the attacker wishes to gain access to a secure facility controlled by a whitelist of people who have
access.

In this scenario, we may conclude that it is very easy or very difficult for an attacker to mount a correlated
input attack depending on a few different factors.
First, if we believe that the attacker must probe the system for errors by physically arriving at the facility
and attempting to enter wearing different glasses, it seems difficult to mount a correlated input attack.
The attacker may be arrested by the guard if the face recognition system correctly rejects them as not belonging
to the whitelist. An attacker who arrives repeatedly wearing different glasses each time may be even more likely
to be arrested.
This means that there is little opportunity to probe the system for errors and also a high cost to mounting
failed attacks.
Under such a scenario, we may actually be interested in the error rate on adversarial modifications of naturally
occurring data (i.e., we may be interested in the percentage of people who can appear to be on the whitelist
by wearing adversarial glasses). If this error rate is low, attackers will be discouraged from mounting exploratory
attacks.
If, on the other hand, we believe that it is easy to transfer adversarial examples from other models, then
attackers may be able to build their own model or ensemble of models \citep{szegedy2013intriguing,liu2016delving},
and find a person for whom some set of glasses will reliably get them into the facility.
Or, if the face recognition system used by the facility is commercially available, the attackers could buy their
own copy, test it for vulnerabilities offline, and then mount a live attack after finding a reliable vulnerability.
In either of these cases, the attacker now has a high ability to mount a correlated input attack.
However, we may still find some use in studying the error rate on adversarial modifications of naturally sampled
data.
This error rate is essentially the percentasge of the general population who can be used to mount a reliable attack,
given the correct glasses. If the error rate of the system is small enough, it may be difficult for the attacking
organization to find and recruit an individual who has both the right face and the skills to carry out the adversarial
mission.
This is overall merely a mitigating factor and not a complete defense though.
We have seen that in this example, the ability of the attacker to perform offline screening determines whether
defenses that reduce the error rate on adversarial perturbations of naturally occurring data are of some limited uses
(when the attacker can perform screening) or are of relatively high value as a deterrent (when the attacker must
perform exploratory attacks live, and faces a high risk of arrest if the error rate is low).

As another example, consider a detector of synthetic media (``DeepFakes'' as discussed by \citep{chesney2018deep}).
A game theoretic analysis of the fake-vs-fake detectors competition \citep{goodfellow2014generative} suggests that, given enough computation and enough data,
the Nash equilibrium is for the fakes to come from the same distribution as real data, forcing the fake detector
to perform no better than chance.
This problem can be avoided in the short term while the fakes and fake detectors are in an ``arms race'' approaching
the Nash equilibrium.
The long run is beyond the scope of this article (fake detectors could perform better if given access to signals
other than the content of the media itself, tools such as cryptographic signing of real media may be more useful
than fake detectors, etc.).
As an exercise in threat modeling, consider how to evaluate a short term fake detector.
Even a fake detector that performs well on randomly sampled data (accuracy of 99\% on a collection of real images
and a collection of fake images from some generative model) could perform quite poorly in practice.
The attacker only needs to find one fake image that bypasses the detector, and then this image can be deployed
widely. In fact, with the fake detector's stamp of approval, the image would be even more credible than if
no fake detector existed. In this scenario, correlated data attacks are
much more likely to be feasible because the attacker can
upload multiple candidate fake images from multiple anonymous accounts and observe which are flagged as fake
with impunity.
Moreover, it is not particularly important for the attacker to be able to choose exactly which image results
in a mistake.
If the motive of the attacker is to cause political damage to a particular political cause, the set of images
damaging to that cause is often quite large.
It is very different from the previous face recognition example, where it must be possible to cause a mistake
using the face of one of the infiltrators as the starting point for the attack.
In this hypothetical scenario, it is particularly important to build models that are robust to correlated data
attacks.

\section{Limitations of current defenses}

Most current defenses are intended to mitigate problems caused by adversarial
distribution shift, but not problems caused by adversarial correlation of
test-time examples.

Most strategies for mitigating distribution shift can still be highly exploited
by an adversary who can impose correlation of test time examples.

One popular family of strategies for mitigating distribution shift is
{\em adversarial training} \citep{szegedy2013intriguing,Goodfellow-2014-adversarial,madry2017towards}.
Current state of the art adversarially trained models are all still deterministic,
so if they have non-zero error rate on naturally occurring data, an attacker
can find a single mistake and then repeat it in order to obtain essentially a 100\%
attack success on correlated attack data.
While adversarial training has thus been state of the art on many expectimax research
benchmarks (expectation over I.I.D. test examples, max over error induced by adversarial
perturbations), it is not useful for resisting true worst case attacks that occur in practice.
A similar criticism applies to all other current defense techniques designed to
find a fixed decision boundary that reduces an expectimax metric, including certified defenses
(e.g. \citet{wong2017provable,raghunathan2018certified,dvijotham2018training,cohen2019certified}).
This is not to criticize adversarial training or certified defenses against perturbations,
because they are useful for making algorithms that perform well despite a change in the data-generating
distribution.
The eventual solution will need to address both changes in the data-generating distribution
and relaxation of the assumption that test-time examples are generated independently.

\section{Hello world example}

As a brief example, let's walk through a few hypothetical attacks that use correlated data,
and a few hypothetical defenses.
All of the attacks will be variations of the ``test set attack'' described by \citet{gilmer2018motivating}.
The defenses will be variations on three themes---improved supervised learning performance,
generic classifiers using a fixed distribution,
and test set memorization.
All of these are ``hello world'' examples of very weak attacks and very weak defenses.
My intention in writing this article is to encourage others to develop better defenses
in this space (and insofar as better attacks are necessary to study better defenses,
also to develop better attacks).

The test set attack basically consists of presenting examples from the test set i.i.d.
to the classifier until the classifier makes a mistake. That mistake is then repeated
indefinitely.
There are obviously many ways to formalize this in practice, allowing us to pay attention
to factors like rate-limiting access to the classifier, the classifier accepting batches
of data vs individual examples, how the classifier performs on data other than the data
provided by a particular adversary, how long the classifier will stay live before being
replaced with a different one, etc. Here I will focus on a few particularly simple cases.

Suppose that the adversary is given a finite number of opportunities to attack the classifier,
and thus essentially just presents a test set containing $\mte$ examples.

Consider an untargeted attack in which the goal of the adversary is simply to cause misclassification.
If the classifier has error rate $r$, then the expected number of trials to find a mistake is
$\frac{1}{r}$. For the remaining test examples, the attacker simply repeats this mistake.
For example, on CIFAR-10, an attacker allowed to present the same number of examples as in the CIFAR-10
test set (10,000), when attacking a classifier with $r=.02$, would need on average 50 examples to find
a mistake, and then would cause mistakes on the remaining 9950 examples, for a total error rate of
$99.51\%$.
This is thus a fairly strong attack against an undefended model.
The attack could be made stronger by using black box adversarial examples to increase $r$ beyond the
rate on naturally occurring data.

How do the defenses fare?

{\em Adversarial training, etc}:
Existing adversarial robustness approaches such as adversarial training generally do not reduce $r$
on naturally occurring data, so they do not help at all. In fact, many of them increase $r$ and would
slightly hurt performance.

{\em Better supervised learning}:
\citet{gilmer2018motivating} suggest focusing on traditional supervised learning metrics such as the
error rate on naturally occurring data, and also suggest reducing the total volume of input space
that results in an error.
Overall the error rate under the test set attack obtained by a classifier with error rate $r$ on naturally occurring data is
\[
  \tilde{r} = \frac{\mte - \frac{1}{r}} {\mte}.
\]
For any $r$ greater than zero, this error rate approaches $1$ as $\mte$ approaches infinity.
In other words, if the classifier is deployed for long enough, an attacker can exploit it arbitrarily badly.
Or if we consider a relatively short finite deployment, for our hypothetical CIFAR-10 classifier to achieve
an $\tilde{r}$ of .05, it would need to reach an $r$ of roughly $10^{-4}$.
On CIFAR-10 I do not know of a good way to estimate the Bayes error rate, but it seems reasonable to believe
that for many tasks the Bayes error rate is well above $10^{-4}$, so for these tasks no deterministic classifier could be
reasonably secure.

{\em Stochastic models}
If a deterministic model performs badly, what about a stochastic model?
Stochastic models have been proposed as defenses against perturbation-based expectimax adversarial examples \citep{feinman2017detecting}.
While the specific defenses proposed so far are known to be mostly broken in most cases \citep{carlini2017adversarial}
we can still think about how well stochasticity in general can perform as a defense against the test set attack.
Suppose that rather than learning a deterministic classification function mapping example $\vx$ to class $y$,
the model uses a fixed classification distribution, $\pmodel(y \mid \vx)$.
These models are still quite vulnerable.
Suppose the attacker manages to find a point $\vx$ such that, when sampling class estimates from the model,
the model has error rate $r_{\vx}$.
Then, asymptotically as $\mte$ approaches infinity, the model under attack using $\vx$ has error rate $r_{\vx}$.
If there exists even a single point where the true class is not $\argmax_y \pmodel(y \mid \vx)$,
then $r_{\vx} > 0.5$.
Overall, the stochasticity can make the attack more complicated: the attacker should not just repeat the first
$\vx$ that results in a mistake because the mistake may not be repeatable, the attacker has to maintain an
estimate of $r_{\vx}$ for each $\vx$ the attacker has tried, and the attacker faces an exploration-vs-exploitation
tradeoff (if the attacker has found a $\vx$ with $r_\vx > 0.5$, should they repeat it indefinitely, or search
for another input with an even higher error rate?).
In other words, the attacker must upgrade from the simplest version of the test set attack to some kind of
rate-tracking, rate-optimizing test set attack.
Asymptotically though, the stochastic defense can bring the error rate down only as low as $50\%$, unless the
baseline $r$ has literally reached $0$.

{\em Abstention}: suppose that a model refuses to classify some examples, either determinstically, or by
sometimes sampling an ``abstain'' class from a fixed $\pmodel(y \mid \vx)$. (Depending on the application / threat model,
it may or may not be feasible to allow the model to abstain)
This does not change the above analysis
very much. The model will presumably abstain on many mistakes and thus reduce $r$, if we define $r$ to be the
rate at which the model makes an incorrect prediction rather than abstaining or making a correct prediction.
However as long as we do not reduce $r$ to literally $0$ the above analysis still applies.

{\em Dynamic modeling}: I argue that the problem is using a fixed model $\pmodel(y \mid \vx)$, regardless of whether
this model is deterministic, makes use of probabilities, or is able to abstain on some inputs.
Instead, I believe that the model must become a moving target that cannot be repeatedly exploited using the same
input.
The concept of a moving target defense has precedent in traditional computer security
\citep{jajodia2011moving}.
As a ``hello world'' example showing that improvements are possible using a dynamic strategy, consider
{\em the memorization defense}: the classifier memorizes all previously seen $\vx$ inputs and abstains every time
a repeated $\vx$ is presented. The test set attack thus obtains $\tilde{r} \leq r$, depending on whether the attacker
runs unique test examples trying to explore for a new vulnerability, or whether the attacker runs multiple memorized
points that are abstained on. Note that the abstention rate on adversarial inputs approaches $1$ asymptotically
if the attacker prioritizes exploitation over exploitation.
In many domains, a high abstention rate on adversarial data is acceptable (e.g., a spam detector
that abstains in the presence of an adversary can be acceptable if messages resulting in abstention are treated as spam).
The main downside of this approach is that it can have high abstention rates on non-adversarial data.
On i.i.d. test benchmark datasets containing high dimensional data, the abstention rate will usually be zero,
but in real-world applications where many users send the same query (e.g., one legitimate e-mail is sent to two users,
who then each query the server to find out whether it is spam) the abstention rate of this approach could be significant.
The test set attack and memorization defense could also clearly be extended in an arms race, with the attack being
extended to add noise or other minor variations to avoid the detection of duplicates, and the defense being extended
to reject approximate matches rather than exact matches. The purpose of this article is not to analyze how that arms
race would play out, but just to show that there exists a simple attack for which there exists a dynamic defense
better than all fixed defenses.

Finally, let's consider what happens with targeted, rather than untargeted attacks. For a targeted attack, the
attacker wishes to hit a specific target class, rather than merely causing any mistake.
For most of the analysis above, this just means we use a different value of $r$.
When calculating $r$, we now need to measure the rate at which inputs are misclassified specifically as the
attacker's target class, not just the rate at which they are misclassified.
One rough way to model what happens in this case is just to divide our earlier value for $r$ by $k$ (the number of classes).
In domains where $k$ is very large (e.g. speech recognition, where $k$ is combinatorially large because the model outputs
a complete sentence containing many words / characters) the test set attack is no longer very effective even against
undefended models.
In domains where $k$ is medium (e.g. CIFAR-10, where $k$ is 10, or ImageNet, where $k$ is 1000) it is still feasible
to find targeted attacks within standard benchmark test sets.
The analysis of a stochastic model $\pmodel(y \mid \vx)$ is also different in the targeted case, because
the existence of a point $\vx$ that is argmax-classified as the target class does not imply an attack success rate
of $\frac{1}{2}$ anymore, but only an attack success rate of $\frac{1}{k}$.
Overall, this analysis suggests that for targeted attacks, our ``hello world'' test set attack is not nearly as interesting
as in the untargeted setting.
Finally, for the memorization defense, one interesting observation is that the model does not need to be able to abstain
in order to reduce the attack success rate.
Rather than abstaining on memorized examples, the model can return a class uniformly at random, reducing the error rate
to $\frac{1}{k}$.
The memorization defense without abstention thus reduces $\tilde{r}$ of an arbitrary stochastic or deterministic model
to match the $\tilde{r}$ of the best possible stochastic model for a fixed argmax-classification $r$.

So far the defenses here have been discussed primarily in a black box setting, where the attacker is able to send inputs
and observe outputs, but does not have a full specification of the model.
In the white box setting, the attacker could perform exploratory screening on their own copy of the model.
When mounting the real attack, they could thus obtain the asymptotic error rate immediately, rather than ramping up to
it after an initial exploratory phase.
For deterministic or fixed stochastic models, this asymptotic error rate can be obtained using a single adversarial
example.
For a dynamic model using the memorization defense, a white box threat model is enough to break the defense and
obtain worse results than the asymptotic black box setting.
During the offline screening phase, if the attacker is able to find $\mte$ unique mistaken points, then during the
online attack phase the attacker can present each of these points to obtain an error rate of $100\%$.
In the future, I hope that a more sophisticated dynamic defense might be able to succeeded in a ``white fading to black box'' case
where the attacker has a full description of the model at some point in time, but the model's dynamic updates are
unpredictable enough that the attacker rapidly loses knowledge of the model as the model is updated.

Both the variations on the test set attack presented here and the memorization defense presented here are intended
as ``hello world'' examples to show that the test set attack, though trivial, can seriously compromise existing defenses,
and that the memorization defense, though impractical for real-world use, can reach levels of robustness not reachable
by a fixed model. I hope that these ``hello world'' examples will pave the way for more practical defenses against
correlated data attacks in the future.

\section{Combining defense strategies}

While I am fairly confident that dynamic model behavior will be a necessary
component of an eventual successful defense strategy, I doubt that dynamic
model behavior alone will be sufficient.
I think it is most likely that the performance of a model under attack is
something like an ``AND'' function: many factors need to be handled correctly
for the defense to work, and the performance under attack will be near zero
if any one of those factors is absent.
Thus, introducing a single component, such as dynamic model behavior, may
result in no measurable benefit until some other required synergistic component
is introduced.

I speculate that the following may also be important mechanisms to combine with
dynamic model behavior.

\subsection{Confidence thresholding / abstention mechanisms / detection mechanisms}

In the ``hello world'' memorization defense above, I incorporated an abstention
mechanism in order to achieve a good defense in the untargeted setting.
This was not necessary in the targeted setting.

I suspect that in many practical settings, abstention will be an important mechanism
to include in dynamic defenses. 
In the ``hello world'' defense, the model abstained on memorized examples that were
previously presented. I suspect that other defenses will need to abstain using other
criteria.

Suppose that we make a dynamic model that constantly moves its decision boundary,
so that attackers cannot predict exactly what it will do. Its decision boundary needs to
remain somewhat near the true decision boundary or it will not perform well on
naturally occurring data.
For points near the decision boundary, the dynamic model is thus forced to behave more
deterministically.
One solution is to abstain on such points.
This is just one example of an abstention criterion different from the memorization criterion.
Other researchers are also exploring abstention mechanisms in other contexts
\citep{carlini2017adversarial,papernot2018deep,goodfellow2019evaluation,brown_contest}.
In general, I think abstention using a variety of criteria may be necessary to combine with
dynamic modeling.

  \subsection{Active learning}
  Another strong candidate is active learning. When a model detects potentially
  troubling activity at test time it could request ground truth labels for some
  of these examples.
  One limitation of the memorization defense is that, while it can detect repeated examples
  as potentially troubling activity, it does not have a mechanism for determining the correct
  way to label them.

  \subsection{Limiting access to the model}

  For many applications, legitimate users do not need to run many queries.
  Users mounting the test set attack or similar black box attacks need to run relatively
  large numbers of queries (e.g. $\frac{1}{r}$ for the test set attack).

  Dynamic models that change during their deployment phase may have some limit on how
  quickly they can change without compromising performance on naturally occurring data.
  Limiting access to the model, for example, rate-limiting the number of queries a user
  can make, may be an important mechanism to make sure that attackers cannot find
  vulnerabilities as quickly as dynamic updates to the model remove them.

\section{Conclusion}

Based on the above line of reasoning, I believe it is important for more of the
machine learning research community to study defenses against attackers who use
correlated attacks at test time, such as finding a single mistake and repeatedly
exploiting it.
I do not have a complete solution in mind, but I believe that any fixed model,
even a fixed model that incorporates stochasticity, will always be a ``sitting
duck'' that can be reliably broken as soon the attacker has found a weak point.
Because of this, I believe it will be necessary to develop dynamic models that
change their decision function continually during deployment.
I suspect it will also be necessary to combine such behavior with other defense mechanisms,
such as confidence thresholding, active learning, and limiting access to the model.
I call on the machine learning research community to help make such effective defenses a reality.

\subsubsection*{Acknowledgments}

Many thanks to Nicolas Papernot for helpful discussion of drafts of this article.

\bibliography{iclr2019_conference}
\bibliographystyle{iclr2019_conference}

\appendix

\end{document}